\documentclass[conference]{IEEEtran}
\usepackage{cite}
\usepackage{amsmath,amssymb,amsfonts}
\usepackage{graphicx}
\usepackage{textcomp}
\usepackage{xcolor}
\def\BibTeX{{\rm B\kern-.05em{\sc i\kern-.025em b}\kern-.08em
    T\kern-.1667em\lower.7ex\hbox{E}\kern-.125emX}}

\usepackage{algorithmic}
\usepackage{algorithm}
\newcommand{\calE}{{\mathcal E}}
\newcommand{\calF}{{\mathcal F}}
\newcommand{\calG}{{\mathcal G}}
\newcommand{\calN}{{\mathcal N}}
\newcommand{\calX}{{\mathcal X}}
\newcommand{\dd}{\mathrm{d}}
\newcommand{\EE}{\mathrm{E}}
\newcommand{\Eqref}[1]{Eq. \eqref{#1}}
\newcommand{\fhat}{\widehat{f}}
\newcommand{\LPi}{L^2} 
\newcommand{\Var}{\mathrm{Var}}

\newtheorem{Assumption}{Assumption}
\newtheorem{Definition}{Definition}
\newtheorem{Proposition}{Proposition}
\newtheorem{Theorem}{Theorem}

\newcommand{\myfootnote}[1]{
  \renewcommand{\thefootnote}{}
  \footnotetext{\hspace{-16.5pt}#1}
  \renewcommand{\thefootnote}{\arabic{footnote}}
}

\begin{document}

\title{Scalable Federated Learning for Clients with Different Input Image Sizes and Numbers of Output Categories}

\author{\IEEEauthorblockN{1\textsuperscript{st} Shuhei Nitta}
\IEEEauthorblockA{\textit{Corporate R$\&$D Center} \\
\textit{Toshiba Corporation}\\
Kawasaki, Japan \\
shuhei.nitta@toshiba.co.jp}
\and
\IEEEauthorblockN{2\textsuperscript{nd} Taiji Suzuki}
\IEEEauthorblockA{\textit{Graduate School of Information Science and Technology, The University of Tokyo} \\
\textit{Center for Advanced Intelligence Project, RIKEN}\\
Tokyo, Japan \\
taiji@mist.i.u-tokyo.ac.jp}
\and
\IEEEauthorblockN{3\textsuperscript{rd} Albert Rodr\'{i}guez Mulet}
\IEEEauthorblockA{\textit{Corporate R$\&$D Center} \\
\textit{Toshiba Corporation}\\
Kawasaki, Japan \\
albert1.rodriguezmulet@toshiba.co.jp}
\and
\IEEEauthorblockN{4\textsuperscript{th} Atsushi Yaguchi}
\IEEEauthorblockA{\textit{Corporate R$\&$D Center} \\
\textit{Toshiba Corporation}\\
Kawasaki, Japan \\
atsushi.yaguchi@toshiba.co.jp}
\and
\IEEEauthorblockN{5\textsuperscript{th} Ryusuke Hirai}
\IEEEauthorblockA{\textit{Corporate R$\&$D Center} \\
\textit{Toshiba Corporation}\\
Kawasaki, Japan \\
ryusuke.hirai@toshiba.co.jp}
}

\maketitle

\myfootnote{\copyright 2023 IEEE.}

\begin{abstract}
Federated learning is a privacy-preserving training method which consists of training from a plurality of clients but without sharing their confidential data. 
However, previous work on federated learning do not explore suitable neural network architectures for clients with different input images sizes and different numbers of output categories. 
In this paper, 
we propose an effective federated learning method named ScalableFL, 
where the depths and widths of the local models for each client are adjusted according to the clients' input image size and the numbers of output categories.
In addition, 
we provide a new bound for the generalization gap of federated learning. 
In particular, this bound helps to explain the effectiveness of our scalable neural network approach.
We demonstrate the effectiveness of ScalableFL in several heterogeneous client settings for both image classification and object detection tasks.
\end{abstract}


\section{Introduction}
\label{sec:introduction}
Deep neural networks have achieved impressive output for various machine learning tasks such as image classification, object detection, and machine translation \cite{LeCun2015-DL}. 
In general, deep neural networks require a large amount and variety of training data in order to achieve good performance. 
One could think of sharing similar data from different clients in order to increase the amount of training data, 
however this is unviable due to issues in privacy or confidentiality. 

Federated learning \cite{Konecny2016-FLorg} is a privacy-preserving training method that trains from a large number of clients by aggregating locally updated model parameters instead of the local training data. 
This learning method gives us a clear advantage in terms of privacy oriented industrial applications. 
For example, 
it enables neural network to train for anomaly detection or localization system 
across multiple 
manufacturing factories. 
In practice, the input image sizes and numbers of output categories may differ between clients, 
thus making it difficult to run federated learning using a common neural network architecture suitable for all clients. 

\begin{figure}[ht]
  \begin{center}
    \centerline{\includegraphics[width=\columnwidth]{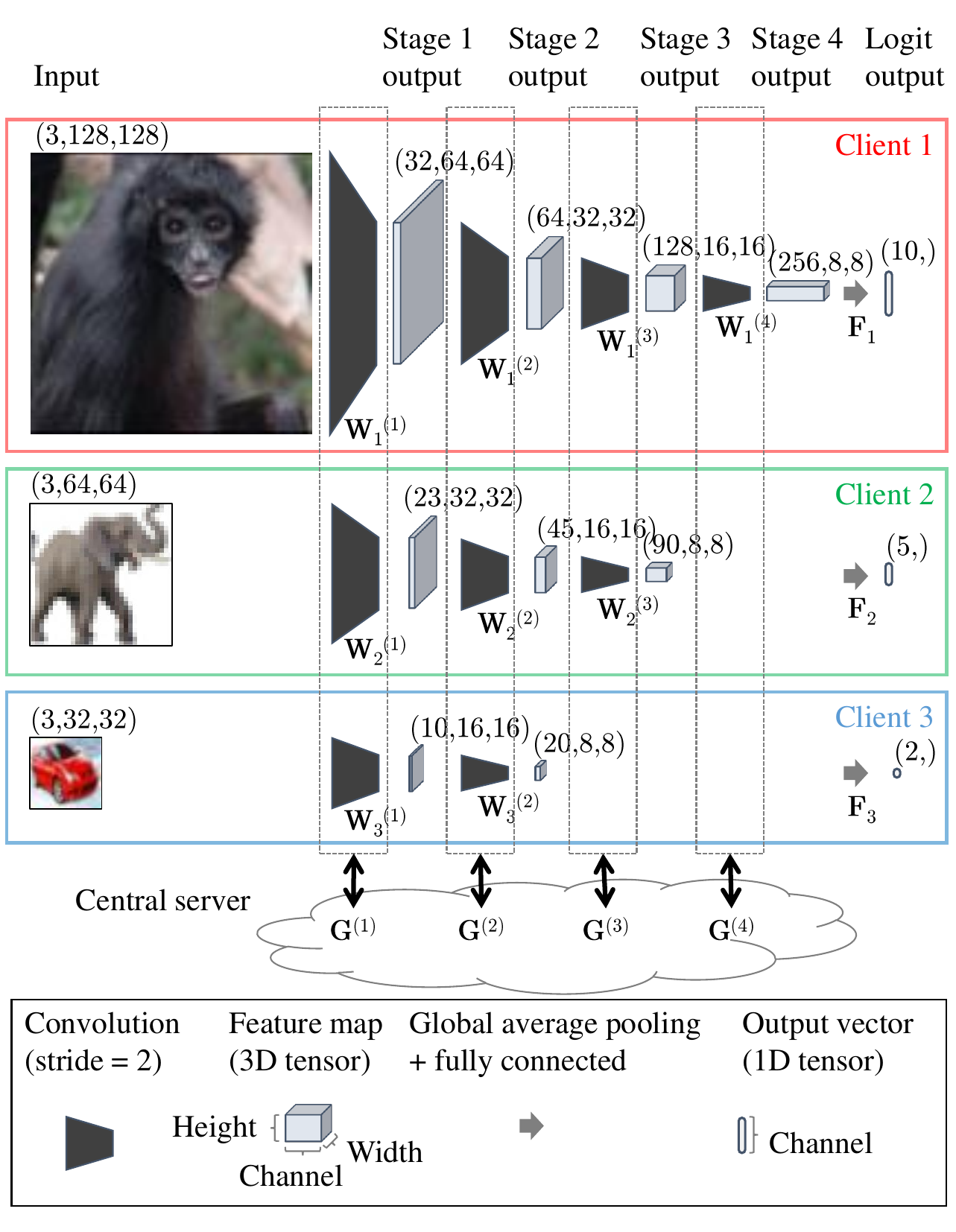}}
    \caption{Example of ScalableFL experiment on tiny image datasets. 
    We assume three clients with different input image sizes and numbers of output categories. 
    Each client uses an effective local model architecture whose depth and width are adjusted 
    according to the input image size and number of output categories. 
    }
    \label{fig:overview}
  \end{center}
\end{figure}

To address this problem, 
we propose an effective federated learning method named \emph{ScalableFL}, 
which uses local models whose depths and widths are adjusted according to the input image sizes and numbers of output categories of the clients, as shown in Fig.  \ref{fig:overview}. 

The contributions of this paper are as follow.
\begin{itemize}
  \item{We introduce an effective method of designing local model architectures whose depths and widths are adjusted according to the input image sizes and numbers of output categories of the client. 
        To the best of our knowledge, our proposed method is the first federated learning approach that trains local models with different depths and widths from a single global model. }
  \item{We provide a new bound for the generalization gap of federated learning. 
        According to the new bound, federated learning using a scalable neural network architecture, including the proposed method, 
        can be interpreted as an effective method that suppresses the upper bound of the generalization gap. }
  \item{The experimental results for image classification and object detection demonstrate that our proposed method is effective for heterogeneous clients with different image sizes and numbers of classes. 
        In particular, our results show significant practical improvements compared with HeteroFL \cite{Diao2021-HeteroFL}, which can train heterogeneous local models (adjusting only the widths).}
\end{itemize}

\section{Related Works}
\label{sec:previous}
Major challenges for federated learning are 
model aggregation \cite{McMahanMRHA2017-FedAvg}, 
privacy preservation \cite{Geyer2017-privacy}, 
communication efficiency \cite{Konecny2016-comm}, and 
personalization.
This paper focuses on personalization, which 
is the adaptation of the model to the specific data properties of each client. 
In particular, we consider an effective federated learning method for heterogeneous clients with different input image sizes and numbers of output categories. 
A variety of personalized federated learning methods have been proposed in order to tackle the challenge of data heterogeneity. 

\paragraph{Local Batch Normalization Layers.} 
FedBN \cite{Li2021-FedBN} uses local batch normalization layers to reduce the feature shift of each client, but this method does not take into account the case where the input image sizes and numbers of output categories differ between clients. 

\paragraph{Local Output Layers.} 
FedPer \cite{Arivazhagan2019-FedPer} 
is a layer personalization approach that combats the adverse effects of the statistical data heterogeneity of each client. 
FedRep \cite{Collins2021-FedRep} is an extension of FedPer, 
providing convergence analysis and proposing to increase the number of local updates for personalized layers.
Furthermore, PartialFed \cite{Sun2021-PartialFed} was proposed, which manually (PartialFed-Fix) or automatically (PartialFed-Adaptive) selects personalized model parameters 
by assuming that the clients have cross-domain characteristics. 
Although these methods can be applied to clients with different numbers of output categories, these methods do not consider the case of different input image sizes. 

\paragraph{Adjusting Local Model Widths.}
HeteroFL \cite{Diao2021-HeteroFL}, 
 FjORD \cite{Horvath2021-FjORD},
 and FedResCue \cite{Zhu2022-FedResCuE}
allow local models to have architectures different from the global model 
in order to adjust them to clients with different necessities, such as different computation and communication capabilities. 
These methods are similar to our approach in the sense that they use different local model architectures.
However, since these methods adjust only the local model width (number of hidden channels), they are not effective against differences in input image size. 
In addition, FjORD and FedResCue require additional computation in order to perform local prunable and self-distillation training. 

\paragraph{Knowledge Distillation.}
Knowledge distillation-based approaches such as QuPeD \cite{Ozkara2021-QuPed} and KT-pFL \cite{Zhang2021-KT-pFL} 
have similar motivations to ours, but require additional computation, communication, or public training datasets. 

\subsubsection{Proposed Method}
In general, as the number of categories increases, the complexity of the task also increases, requiring more feature dimensions.
Also, the larger the image size, the greater the number of suitable layers to ensure that it covers the entire relevant image region. 
Motivated by this observation, we introduce an effective method for designing local model architectures whose depths and widths are adjusted according to the input image sizes and numbers of output categories of the clients. 
To the best of our knowledge, our proposed method is the first federated learning approach that trains local models with different depth and width from a single global model.

The comparison of the personalized federated learning methods is summarized in Table \ref{tab:comparison}. 
FedRep, PartialFed-Adaptive, FjORD, FedResCue, and Knowledge Distillation-based approaches are not included in Table \ref{tab:comparison}, 
because these methods are outside of the scope of the target personalized federated learning method of this paper, 
which is to learn a single global model with slight modifications compared to the local models without additional calculation costs. 

\begin{table*}[t]
  \centering
  \caption{Comparison summary of personalized federated learning methods from a single global model.}
  \label{tab:comparison}
  \begin{tabular}{l|p{1.5cm}p{2cm}p{2.5cm}p{2cm}p{2cm}}
    \hline
                                     & FedBN    & FedPer   & PartialFed-Fix & HeteroFL & ScalableFL \\
                                     & \cite{Li2021-FedBN} 
                                                & \cite{Arivazhagan2019-FedPer} 
                                                           & \cite{Sun2021-PartialFed} 
                                                                            & \cite{Diao2021-HeteroFL} 
                                                                                       &            \\
    \hline
    Local batch normalization layers & $\surd$  & $\times$ & $\surd$        & $\surd$  & $\surd$    \\
    Local output layers              & $\times$ & $\surd$  & $\surd$        & $\times$ & $\surd$    \\
    Adjusting local model widths     & $\times$ & $\times$ & $\times$       & $\surd$  & $\surd$    \\
    Adjusting local model depths     & $\times$ & $\times$ & $\times$       & $\times$ & $\surd$    \\
    \hline
  \end{tabular}
\end{table*}

\section{Methods}
\label{sec:methods}
As we have mentioned in the introduction, 
the main limitations of the existing methods can be summarized in the following two issues: 
(1) since the width of the local model cannot be adjusted according to the number of output categories of the client, 
it is difficult to handle cases where the numbers of output categories are significantly different, and 
(2) since the depths of the local models are not adjusted to the input image sizes of the clients accordingly, 
their capacities are limited or the layers become redundant.
We resolve these two issues by proposing new techniques on the local model width and depth, respectively. 
These techniques can adaptively select the local model structure depending on the task complexity. 
In addition to these techniques, we also adopt local batch normalization and output layers to absorb the difference of the local training sets and tasks. 
The details of the method for model architecture design and optimization are described below. 

\subsection{Architecture Design}
In this section, we formally describe our model architecture design.
Let $M$ be the number of clients. 
For the $j$-th client ($j = 1,..., M)$, 
let $H_j$ be the height (=width) of its input image, and
$K_j$ be the number of output categories. 
The $j$-th client has $c_j$ convolutional layers with kernel size $(3, 3)$, padding $(1, 1)$ and stride $(2, 2)$. 
We represent by ${\rm \bf W}^{(l)}_j \in \mathbb{R}^{a^{(l)}_j \times b^{(l)}_j \times 3 \times 3}$
the parameters of the $l$-th convolutional layer of the $j$-th client, 
where $a^{(l)}_i$ and $b^{(l)}_j$ are the number of input and output channels, respectively. 
In the case of the example in Fig.  \ref{fig:overview}, we have 
$M = 3$, $(H_1, H_2, H_3) = (128, 64, 32)$, and $(K_1, K_2, K_3) = (10, 5, 2)$. 
How to determine the $a^{(l)}_j$, $b^{(l)}_j$, and $c_j$ is described as follows.

\paragraph{Local Model Widths.}
First, we describe the local model widths for each client.
In order to increase the number of channels of feature maps according to the number of output categories,
we introduce a simple hidden channel ratio $\kappa _j$ for the $j$-th client as follows, 
$$
   \kappa _j = \log_{10}(K_j) / \log_{10}(K_0), 
$$
where, $K_0$ is the base number of categories. 
By using the hidden channel ratio, the number of input and output channels are calculated as follows, 
$$
  a^{(l)}_j = \kappa _j a^{(l)}_0, 
  b^{(l)}_j = \kappa _j b^{(l)}_0,
$$
where, 
$a^{(l)}_0$ and $b^{(l)}_0$ are the base number of hidden input and output channels, respectively. 
In the case of the example in Fig.  \ref{fig:overview}, 
we set 
$K_0 = 10$ as the base number of categories and 
$(b^{(1)}_0, b^{(2)}_0, b^{(3)}_0, b^{(4)}_0) = (32, 64, 128, 256)$ as the base number of hidden output channels. 

\paragraph{Local Model Depths.}
Second, we explain the local model depths for each client.
In order to adjust the depths, 
the number of convolutional layers $c_j$ of the $j$-th client is calculated based on the input image height $H_j$ as follows, 
$$
  c_j = \lceil \log_2 H_j / H_0  \rceil, 
$$
where, $H_0$ is the base size of the last feature map. 
Specifically, based on the general experimental conditions, we use $H_0 = 8$ and $4$ for image classification task and object detection task, respectively.
In the case of the example in Fig.  \ref{fig:overview}, we have 
$(c_1, c_2, c_3) = (4, 3, 2)$. 

\paragraph{Local Batch Normalization and Output Layers.}
In addition, 
in order to personalize the training for each of the heterogeneous clients with different input image sizes and numbers of output categories, 
we use private batch normalization layers as in FedBN and we place a head layer ${\rm \bf F}_j \in \mathbb{R}^{b^{(c_j)}_j \times K_j}$ 
after the feature extraction and global average pooling as in FedPer.

\paragraph{Global Model Architecture.}
Next, we present the global model stored in the server.
The global model architecture is conformed by $c_g = \max(\{c_j\}^{M}_{j=1})$ convolutional layers with kernel size $(3, 3)$, 
each with parameters ${\rm \bf G}^{(l)} \in \mathbb{R}^{a^{(l)}_g \times b^{(l)}_g \times 3 \times 3}$, 
where, $a^{(l)}_g = \max(\{a^{(l)}_j\}^{M}_{j=1})$ and $b^{(l)}_g = \max(\{b^{(l)}_j\}^{M}_{j=1})$. 
The local model parameters ${\rm \bf W}^{(l)}_j$ are a subset of the corresponding global model parameters, 
and can be expressed in a similar fashion to slice rule of Pytorch \cite{pytorch2019} as 
${\rm \bf W}^{(l)}_j = {\rm \bf G}^{(l)}$ \texttt{[ :} $b^{(l)}_j$ \texttt{, :} $a^{(l)}_j$ \texttt{, :, :]}.
In the case of the example in Fig.  \ref{fig:overview}, we have 
$c_g = 4$ and $(b^{(1)}_g, b^{(2)}_g, b^{(3)}_g, b^{(4)}_g) = (32, 64, 128, 256)$. 

\subsubsection{Summary of Model Architectures}
We show that this neural network architecture is indeed effective and thus it can be used with heterogeneous clients as described above 
by increasing the number of hidden channels according to the number of output categories and 
   increasing the number of hidden layers according to the input image size. 
As shown in Table \ref{tab:comparison}, 
HeteroFL uses hidden channel ratios and local batch normalization layers, 
but does not use local output layers and depth adjustable models for the clients. 
In addition, 
the PartialFed-Fix uses local batch normalization and output layers, 
but does not use hidden channel ratios and depth adjustable models for the clients.

Our proposed method can be extended to major convolutional neural network architectures such as 
VGG \cite{Simonyan2015-vgg}, 
ResNet \cite{He2016-resnet,He2016-PreActResNet}, 
MobileNet \cite{Howard2017-mobile1,Sandler2018-mobile2}, 
and EfficientNet \cite{Tan2019-EfficientNet}, and to object detection tasks such as SSD \cite{Liu2016-ssd}. 
In all these cases, the number of convolutional layers $L_i$ varies according to the number of stages 
(number of times the feature map width and height is halved) . 

Note that EfficientNet \cite{Tan2019-EfficientNet} fixes the number of stages and adjusts the model depth by changing the number of layers in each stage
and its architectures are huge and have been optimized only experimentally through an extensive number of experiments on a particular large size dataset.
In contrast, our proposed method adjusts the model depth by changing the number of stages, 
which we believe it makes easier to generalize better to a wider range of possible image input sizes and datasets for 
image classification and object detection using feature maps of each stage. \footnote{
  In this paper we propose a very simple and straightforward yet effective method to determine the widths and depths of the local models. 
  We too acknowledge that this approach needs further exploration but we leave it as future work. 
}

\subsection{Optimization}
The procedure for the optimization of ScalableFL is summarized in Algorithm \ref{alg:scalableFL_opt}. 
In each communication round, 
the learning rate for local model updates is adjusted (line 2), and 
in order to update each model parameter in a well-balanced manner, 
$\xi$\%
of clients are sampled for each client group with the same input image size and number of categories (line 3). 
Let ${\mathcal S}$ be the set of indexes of the selected clients. 
Each selected client $j \in {\mathcal S}$ gets the subset of shared model parameters ${\mathcal W}_j$ 
from the global model (line 5) and updates the local model (line 6).
The specific update method is explained in the next paragraph.
After updating, global model parameters are aggregated using element-wise average of the shared weights of selected local model parameters $\{{\mathcal W}_j\}_{j \in {\mathcal S}}$ (line 9). 
Algorithm \ref{alg:scalableFL_opt} is simlar to FedAvg \cite{McMahanMRHA2017-FedAvg}, 
but line 5 (shared weights are a "subset" of the corresponding global model parameter) and 
line 9 ("element-wise" averaging) are different in order to accommodate local models whose widths and depths are different. 

The local model update method is summarized in Algorithm \ref{alg:scalableFL_local}.
First, the local client creates the local model ${\mathbf \Theta}_j$ from 
the set of shared model parameters ${\mathcal W}_j$, 
the set of local batch normalization parameters ${\mathcal B}_j$, and
the local output layer parameters ${\rm \bf F}_j$ (line 1). 
After the model creation, 
$E$ times of mini batch local training based on the loss function ${\mathcal L}$ of the target task and 
weight decay regularization ${\mathcal R}$, are performed (line 3, 4).
Momentum SGD is used as the optimizer, and the learning rate $\eta$ and weight decay strength $\lambda$ are manually tuned for each experiment.

\begin{algorithm}[tb]
  \caption{ScalableFL optimization}
  \label{alg:scalableFL_opt}
  \begin{algorithmic}[1]
    \REQUIRE
      number of communication rounds $T$, 
      participation rate $\xi$, 
      set of global model parameters $\{{\rm \bf G}^{(l)}\} ^{N_g}_{l = 1}$,
      initial learning rate $\eta$
      
    \FOR{ each communication round to $T$ }
      \STATE{$\eta \leftarrow$ AdjustLearningRate $(\eta)$}
      \STATE{ $\mathcal S \leftarrow $ SelectClients($\xi$)}
      \FOR{ each client $j \in \mathcal S$ {\bf in parallel}}
        \STATE{${\mathcal W}_{j} \leftarrow $ get set of shared weights $(\{{\rm \bf W}^{(l)}_{j} \leftarrow {\rm \bf G}^{(l)}[:b^{(l)}_j, :a^{(l)}_j, :, :] \}^{N_j}_{l = 1})$}
        \STATE{${\mathcal W}_{j} \leftarrow $ LocalUpdate $({\mathcal W}_{j}, \eta)$}
      \ENDFOR
      \FOR{ $l = 1,..., N_g$ }
        \STATE{${\rm \bf G}^{(l)} \leftarrow$ element-wise averaging of $\{{\rm \bf W}^{(l)}_{j}\}_{j \in \mathcal S}$}
      \ENDFOR
    \ENDFOR
  \end{algorithmic}
\end{algorithm}

\begin{algorithm}[tb]
  \caption{LocalUpdate (${\mathcal W}_j, \eta$) for $j$-th client}
  \label{alg:scalableFL_local}
  \begin{algorithmic}[1]
    \REQUIRE
      number of local iterations $E$, 
      local dataset ${\mathcal D}_j$,
      set of local batch normalization parameters ${\mathcal B}_j$,
      local output layer parameters ${\rm \bf F}_j$,
      loss function ${\mathcal L}$,
      weight decay function ${\mathcal R}$ and its coefficient $\lambda$
      
    \STATE{${\mathbf \Theta}_j \leftarrow $ CreateLocalModel $({\mathcal W}_j, {\mathcal B}_j, {\rm \bf F}_j)$}
    \FOR{ $e = 1,..., E$ }
      \STATE{${\mathcal D}_{j,e} \leftarrow$ get next mini batch }
      \STATE{${\mathbf \Theta}_j \leftarrow $ LocalOptimizer $({\mathbf \Theta}_j, \nabla({\mathcal L}({\mathbf \Theta}_j, {\mathcal D}_{j,e}) + \lambda {\mathcal R}({\mathbf \Theta}_j) ),\eta )$}
    \ENDFOR
    \RETURN{${\mathcal W}_j$}
  \end{algorithmic}
\end{algorithm}

\section{Theoretical Analysis}
\label{sec:support}
In this section, we give a generalization error analysis for a rather wide class of federated learning
that justifies the benefit of our scalable approach, and in particular, the benefit of the model sharing method.
For that purpose, we let $\calF$ be the set of functions in our largest (global) model,
and we consider a sequence of submodels $f^{(j)}~(j=1,\dots,M)$ extracted from the full model $f \in \calF$ to solve each local problem $j \in [M] := \{1,\dots,M\}$.
Here, $M$ is the number of tasks each of which corresponds to a local environment.
The set of local models $f^{(j)}$ is denoted by $\calF_j$ and we consider a setting where the local models have a hierarchy like $\calF_1 \subset \calF_2 \subset \dots \subset \calF_M = \calF$.
We let $f^{(0)} = 0$ for convention. 
For each task $m \in [M]$, we are given training data $(z_i^{(m)})_{i=1}^{n_m} = (x_i^{(m)},y_i^{(m)})_{i=1}^{n_m}$ with size $n_m$ where $x_i^{(m)}$ is an input and $y_i^{(m)}$ is the corresponding label (just for mathematical simplicity, we let the spaces of input and output are same across all tasks, but our theory trivially covers the setting where the resolution of input image is different). 
We let $\bar{n}_j := \sum_{m'=j}^M n_{m'}$ (sample size of ``upper level'' tasks) and $n = \sum_{m=1}^M n_{m}$.
Accordingly, we can define the training loss as 
$$
\textstyle
\hat{L}(f) = \frac{1}{\sum_{m=1}^M n_m}\sum_{m=1}^M \sum_{i=1}^{n_m} \ell(f^{(m)},z^{(m)}_i),
$$
for $f \in \calF$
where $f^{(m)}$ is the submodel of $f$ corresponding to the $m$-th task and $\ell$ is a loss function of the model $f^{(m)} \in \calF_m$ on the training data-point $z_i^{(m)}$.
The expected loss (test loss) is also defined as 
$$
\textstyle
L(f) =\sum_{m=1}^M \alpha_m \EE[\ell(f^{(m)},Z^{(m)})], 
$$
where $\alpha_m = \tfrac{n_m}{\sum_{m'=1}^M n_m'}$ and the expectation is taken over the input-output pair 
$Z^{(m)} = (X^{(m)},Y^{(m)})$ for each task.
%
%
%

Let $P_m$ be the distribution of data $Z^{(m)} = (X^{(m)},Y^{(m)})$ for the $m$-th task.
We define the empirical and population $L^2$-norms for each task as 
$$
\|f\|_{n_m}^2 := \frac{1}{n_m}\sum_{i=1}^{n_m} f(x_i^{(m)})^2,~~~\|f\|_{L^2(P_m)}^2 := \EE[f(X^{(m)})^2].
$$
We also define 
$\bar{P}_j := \sum_{m=j}^M \alpha_m P_m/(\sum_{m=j}^M \alpha_m)$, and 
$
\|f\|_{\bar{n}_j}^2 := \frac{1}{\bar{n}_j} \sum_{m=j}^M  \sum_{i=1}^{n_m} f(x_i^{(m)})^2$
and $\|f\|_{L^2(\bar{P}_j)}^2 := \frac{1}{\sum_{m=j}^M \alpha_m }\sum_{m=j}^M \alpha_m \|f\|^2_{L^2(P_m)}$,
which are important to define the {\it local Rademacher complexity} defined below.

Here, we introduce the local Rademacher complexity, which is a typical tool to evaluate the generalization error,  
as $\hat{R}_{j,r}(\calF_j) :=\EE_{\epsilon}\big[\sup\{\frac{1}{\bar{n}_j} \sum_{m=j}^M 
\sum_{i=1}^{n_m} \epsilon_{i,m}(f^{(j)}(x_i^{(m)})-f^{(j-1)}(x_i^{(m)}))
\mid f^{(j)} \in\calF_j, \|f^{(j)}-f^{(j-1)}\|_{\LPi(\bar{P}_j)} \leq r \} \big]$
where 
$r > 0$ is arbitrary and 
$\epsilon_{i,m}~(i=1,\dots n_m, m=j,\dots,M)$ is an i.i.d. Rademacher sequence $(P(\epsilon_{i,m} = 1) = P(\epsilon_{i,m} = -1)= 1/2)$. 
Its expectation with respect to $D_n = \{ (z_i^{(m)})_{i=1}^{n_m}\}_{m=1}^M$ is denoted by 
$
\bar{R}_{j,r}(\calF_j) := \EE_{D_n}[\hat{R}_{j,r}(\calF_j)].
$
Roughly speaking the local Rademacher complexity measures the size of the model with given {\it radius} $r>0$ to bound the generalization error 
\cite{book:Vapnik:1998,mohri2012foundations}. 
To obtain the generalization error bound, we assume the following conditions.
\begin{Assumption}[Lipschitz continuity of loss function]
\label{ass:LipschitzCont}
The loss function $\ell$ is 1-Lipschitz continuous with respect to the function output:
$
|\ell(f, z) - \ell(g, z)| \leq |f(x) - g(x)|~~(\forall z = (x,y), f \in \calF).
$
\end{Assumption}

\begin{Assumption}[Boundedness of the model]
\label{ass:LinftyBound}
The $L^\infty$-norms of all elements in $\calF$ are bounded by $B \geq 1$: $\|f\|_\infty \leq B$
for all $f\in \calF$.
\end{Assumption}

We also assume that the local Rademacher complexity $\bar{R}_{j,r}(\calF_j)$ has a concave shape with respect to $r > 0$:
Suppose that there exists a function $\phi_j:(0,\infty) \to [0,\infty)$ such that 
$$
 \bar{R}_{j,r}(\calF_j) \leq \phi_j(r)~~\text{and}~~\phi_j(2 r) \leq 2 \phi_j(r)~~(\forall r > 0).
$$
This condition is not restrictive, and usual bounds for the local Rademacher complexity satisfy this condition
\cite{IEEEIT:Mendelson:2002,LocalRademacher}.
Using this notation, we define $r_j^* = r_j^*(t)$ as  
\begin{align*}
\textstyle
r_j^*(t) := \inf\Big\{ r>0 ~ \Big| ~ \frac{8\phi_j(r)}{r^2} + B \sqrt{\frac{4t}{r^2 \bar{n}_j}} +  \frac{2B^2t}{r^2 \bar{n}_j} \leq \frac{1}{2}\Big\}.
\end{align*}

Under these conditions, we obtain the following generalization error bound.
\begin{Theorem}

Suppose that the trained network $\fhat \in \calF$ satisfies
$\|\fhat^{(j)} - \fhat^{(j-1)}\|_{\bar{n}_m} \leq \hat{r}_j^2~(\forall j, \forall m \geq j)$ for a fixed $\hat{r}_j > 0$ almost surely. 
Then, for $t > 0$, we let $\dot{r}_j := \sqrt{2(\hat{r}_j^2 + r_j^{*2}(t))}$, then 
under Assumptions \ref{ass:LipschitzCont} and \ref{ass:LinftyBound}, there exists a universal constant $C>0$ such that 
\begin{align*}
& \hat{L}(\fhat) - L(\fhat) \\
  & \leq \textstyle C \sum_{j=1}^M \left(\sum_{m=j}^M \alpha_m\right) \Big[ \bar{R}_{j, \dot{r}_j}(\calF_j) 
\sqrt{\log(\bar{n}_j)} \log(2 \bar{n}_j B) \\
&\textstyle \phantom{\leq}~~+ \dot{r}_j \sqrt{\frac{t + \log(2M)}{\bar{n}_j}} + \frac{1 + B (t +\log(2M)) }{\bar{n}_j}\Big],
\end{align*}
with probability $1 - \exp(-t)$.
\end{Theorem}

This bound can be understood as follows.
Typically, the local Rademacher complexity is given in a form of 
$\bar{R}_{j, \dot{r}_j}(\calF_j) =O\left(\dot{r}_j \sqrt{\frac{\log(N_j)}{\bar{n}_j}}\right)$ where $N_j$ is the {\it covering number} of the model $\calF_j$ that represents the {\it complexity} of the model \cite{pmlr-v65-harvey17a,pmlr-v80-arora18b,Suzuki2020Compression}.
An important point of this evaluation is that the bound is controlled by the radius $\hat{r}_j$ representing the discrepancy between submodels. 
Typically, $\dot{r}_j$ satisfies $\dot{r}_j \sim \hat{r}_j + 1/\sqrt{n_j}$, and our bound can be roughly given as 
$$\textstyle
\sum_{j=1}^M 
\left( \hat{r}_j \left(\tfrac{\bar{n}_j}{n}\right)^{1/2} \sqrt{\frac{\log(N_j)}{n}} + \frac{1}{n}\right).
$$
up to poly-log order. Therefore, if the submodels $f^{(j)}$ and $f^{(j-1)}$ are similar ($\hat{r}_j$ is not large), then the bound becomes tighter.
Indeed, in a situation such that $\sum_{j=1}^M \hat{r}_j =O(1)$ and $N_j$s are same for all $j$, the bound becomes $\sqrt{\frac{\log(N_1)}{n}} + \frac{M}{n}$.
On the other hand, if we train the network independently on the individual task, a naive bound yields 
$$\textstyle
\sum_{m=1}^M \alpha_m \sqrt{\frac{\log(N_m)}{n_m}}
= \sum_{m=1}^M (\tfrac{n_m}{n})^{1/2} \sqrt{\frac{\log(N_m)}{n}}.
$$
Then, if $n_m$ and $N_m$ are uniform for all $m$, then this bound becomes 
$\sqrt{M \frac{\log(N_1)}{n}}$ which is $\sqrt{M}$ times larger than our bound considered above.
This comparison highlights that the strategy to share models across tasks can give better generalization than individually training models.
Our scalable approach realizes this in an efficient way. 
Indeed, our method generates models that fit the data of each task well by choosing appropriate model structures (i.e., yielding a small training error) while it keeps the model size as small as possible (i.e., keeping the local Rademacher complexity small) with small gap of models (i.e., small $\hat{r}_j$). 

There is some related work on generalization error analysis of federated learning.
\cite{9154277,ImprovedarXiv02423} derived an information theoretic bound in which an expectation of the generalization gap is obtained while our bound is a high probability bound. 
Information theoretic quantity in their bound is insightful but does not give a concrete bound. 
On the other hand, our bound successfully gives a bound by explicitly utilizing the distance between submodels and global model.

\begin{table*}[t]
  \centering
  \caption{Experimental results (test accuracy) of image classification task on Imagenet dataset.
           Results of individual learning mean average $\pm$ SD of 5 clients out of the 20 target clients of 1 time training.
           Results of HeteroFL and ScalablFL mean average $\pm$ SD of all target clients of 3 times training.}
  \label{tab:result_imagenet}
  \begin{tabular}{l|rrrr|r}
    \hline
    Client group               & ImageNet-1k            & ImageNet-500           & ImageNet-200           & ImageNet-100           & Average     \\
    \hline
    Individual (5k iters)      & $15.39 \pm 0.21$       & $26.49 \pm 3.36$       & $35.55 \pm 4.58$       & $44.48 \pm 2.30$       & 30.30       \\
    Individual (50k iters)     & $21.49 \pm 0.27$       & $33.03 \pm 4.94$       & $44.13 \pm 4.57$       & $50.87 \pm 1.34$       & 37.13       \\
    Individual (100k iters)    & $20.46 \pm 0.18$       & $31.68 \pm 4.81$       & $42.83 \pm 4.64$       & $50.52 \pm 1.84$       & 36.12       \\
    \hline
    HeteroFL (4 stages)        & $32.42 \pm 0.32$       & $43.26 \pm 5.02$       & $50.03 \pm 3.50$       & $53.69 \pm 2.16$       & 44.85       \\
    HeteroFL (5 stages)        & $34.86 \pm 0.26$       & $43.40 \pm 4.93$       & $45.79 \pm 2.78$       & $48.71 \pm 2.44$       & 43.19       \\
    \hline
    ScalableFL (4 to 5 stages) & ${\bf 35.24} \pm 0.22$ & ${\bf 44.10} \pm 5.10$ & ${\bf 50.99} \pm 3.45$ & ${\bf 54.21} \pm 2.00$ & {\bf 46.14} \\
    \hline  
  \end{tabular}
\end{table*}

\begin{table}[t]
  \centering
  \caption{Experimental results (test mAP) of object detection task on MSCOCO and PascalVOC.
           Results of individual learning mean average $\pm$ SD of 3 clients out of the 16 target clients for MSCOCO or 3 different seeds for PascalVOC.
           Results of HeteroFL and ScalablFL mean average $\pm$ SD of all target clients of 3 times training.}
  \label{tab:result_ssd}
  \begin{tabular}{l|rr}
    \hline
    Client group               & MSCOCO                 & PascalVOC              \\
    \hline
    Individual (5k iters)      & $ 9.18 \pm 0.13$       & $57.71 \pm 0.47$       \\
    Individual (50k iters)     & $15.89 \pm 0.14$       & $65.52 \pm 0.30$       \\
    Individual (100k iters)    & $16.10 \pm 0.11$       & $65.86 \pm 0.37$       \\
    \hline
    HeteroFL (6 stages)        & $15.45 \pm 0.09$       & $66.74 \pm 0.30$       \\
    HeteroFL (7 stages)        & $20.03 \pm 0.11$       & $69.00 \pm 0.29$       \\
    \hline
    ScalableFL (6 to 7 stages) & ${\bf 20.30} \pm 0.09$ & ${\bf 69.98} \pm 0.25$ \\
    \hline  
  \end{tabular}
\end{table}

\section{Results}
\label{sec:results}
We evaluate the performance of our proposed method in three heterogeneous client settings for 
image classification task and object detection task. 
We compare the test accuracy or mAP with that of the individual learning method (training local model from local data only) and HeteroFL \cite{Diao2021-HeteroFL}. 
For individual learning, 
experimental results under various conditions of number of iterations are compared since the optimal number of iterations is often different from federated learning settings. 
Therefore, we tuned the learning rate and weight decay strength for each condition of the number of iterations by grid search. 
Originally, HeteroFL uses federated learning with a slimmable architecture (adjusting only the local model widths) with 
private batch normalization layers. 
In order to apply HeteroFL to our client settings, that is, with different numbers of output categories, 
the output layer is also changed to a private layer like FedPer. 
Therefore, we modify only the hidden channel ratio so that the number of parameters becomes the same as the proposed ScalableFL architectures in each client.
In addition, we report the mean and standard deviation of multiple learning results after the last iteration, and 
all comparison methods are implemented from anew due to our unique experimental conditions.
The details of these model architectures and experimental settings are described in the Appendix. 

\subsection{Image Classification Task on ImageNet}
For the large-scale image classification task, 
we use ImageNet \cite{Deng2009-imagenet} dataset, which has about 1.28M training samples and 1000 categories. 
First, we divide ImageNet dataset into four partitions without duplication, 
and consider it as four different datasets with the same domain. 
For the heterogeneous client settings, 
we split each of these four partitions into 20 clients, 
so that each partition's client consists of 1000, 500, 200, and 100 random classes out of the 1000 classes, 
and the number of training samples of all clients is balanced (about 16k training samples), respectively. 
In this paper, we call the above partitions ImageNet-1k, ImageNet-500, ImageNet-200, and ImageNet-100, respectively.
In addition, we set the input image size of ImageNet-1k, ImageNet-500, ImageNet-200, and ImageNet-100 to 
$256^2, 192^2, 128^2$, and $96^2$ pixels, respectively. 

The base model architecture is ResNet18 \cite{He2016-resnet}, and we use $K_0=1000$. 
Therefore, the number of stages and hidden channel ratio for ImageNet-1k, ImageNet-500, ImageNet-200 and ImageNet-100 are 
$(5, 5, 4, 4)$ and $(1, 0.9, 0.77, 0.67)$, respectively. 
The details of the model architecture and experimental settings are summarized in the Appendix. 

The experimental results are shown in Table \ref{tab:result_imagenet}. 
We evaluate three variations of 5k, 50k, and 100k training iterations with individual learning results, 
and two variations of 4 and 5 stages with HeteroFL learning results. 
In the comparison between ScalableFL and inidividual learning results, 
the performance of individual learning saturates at 50k training iterations. 
On the other hand, ScalableFL outforms the individual learning of ImageNet-1k, ImageNet-500, ImageNet-200 and ImageNet-100 by 
13.8, 12.3, 6.86, and 3.34 points, respectively. 

In the comparison between ScalableFL and HeteroFL learning results, 
ScalableFL can achieve higher performance than HeteroFL in the average of all client results, and
can achieve better performance than HeteroFL using the number of stages performing the best. 
To the best of our knowledge, 
this is the first experiment for ImageNet with 100 category clients. 

\subsection{Object Detection Task on MSCOCO and PascalVOC}
For the object detection task, 
we use MSCOCO \cite{Lin2014-coco} dataset which has 117266 training samples and 80 categories, and 
       PascalVOC \cite{Everingham2015-voc} dataset which has 16551 training samples and 20 categories without background class. 
We split MSCOCO dataset into 16 clients (partitions) and the number of training samples is balanced (about 7.3k training samples). 
Furthermore, we split PascalVOC dataset into 2 clients and the number of training samples is balanced (about 8.3k training samples). 
The input sizes of MSCOCO and PascalVOC are $512^2$ and $256^2$ pixels, respectively. 

The base model architecture is ResNet18 \cite{He2016-resnet} plus two extra stages as the backbone and SSD heads \cite{Liu2016-ssd}, and we use $K_0=80$. 
Therefore, the number of stages and hidden channel ratio of the backbone for MSCOCO and PascalVOC are $(7, 6)$ and $(1.0, 0.68)$, respectivey. 
Details of the model architecture and experimental settings are summarized in the Appendix. 

The results are shown in Table \ref{tab:result_ssd}. 
We evaluate three variations of 5k, 50k, and 100k training iterations with individual learning results, and 
two variations of 6 and 7 stages with HeteroFL learning results. 
In the comparison between ScalableFL and individual learning results of MSCOCO and PascalVOC, 
even in the case of a large number of 100k training iterations, 
ScalableFL outperforms the individual learning by 3.90 and 4.12 points, respectively. 

We showed that, ScalableFL achieves better performance than HeteroFL on all settings. 
We expect that a suitable model architecture (depth) depending on the input image size has a positive effect, 
especially in anchor-type object detection methods. 
To the best of our knowledge, 
this is the first object detection experiment in federated learning with adjustable local model size. 

\section{Conclusions}
\label{sec:conclusions}
In this paper, we proposed an effective federated learning method 
for clients with different input image size and numbers of output categories. 
In order to collaborate with various client settings, 
we introduced an effective method for designing local model architectures 
whose depths and widths are adjusted according to the input image sizes and 
numbers of output categories of the clients. 

In addition, 
we provide a new bound of a generalization gap of federated learning. 
According to the new bound, 
a federated learning using a scalable neural network architecture including our proposed method,
can be interpreted as an effective method that suppresses the upper bound of the generalization gap.

In experiments on an image classification task and an object detection task, 
ScalableFL exhibited favorable performance in several situations. 

\bibliographystyle{ieeetr}
\bibliography{ref}

\clearpage
\onecolumn
\appendix

\begin{center}
{\bf Appendix A: Architecture Details}\\
\end{center}
In Table \ref{tab:arch_imagenet} to Table \ref{tab:arch_ssd_hetero}, we describe the details of the model architecture.

\begin{table*}[t]
  \centering
  \caption{ScalableFL architecture for image classification task on the ImageNet dataset. 
  The base model architecture is post-activation type ResNet18. 
  Downsampling is performed by stage2\_1, stage3\_1, stage4\_1, and stage5\_1 with a stride of 2. 
  In addition the fully connected layers of output\_2 and all batch normalization layers are personalized layers.}
  \label{tab:arch_imagenet}
  \begin{tabular}{c|c||c|c|c|c}
    \hline
    Layer name & Shared     & ImageNet-1k                & ImageNet-500               & ImageNet-200               & ImageNet-100               \\
    \hline
    stage1     & $\surd$    & 7 $\times$ 7, 64, stride 2 & 7 $\times$ 7, 58, stride 2 & 7 $\times$ 7, 50, stride 2 & 7 $\times$ 7, 43, stride 2 \\
    \hline
    stage2\_x  & -          & \multicolumn{4}{c}{ 3 $\times$ 3 max pool, stride 2 } \\
    \cline{2-6}
               & $\surd$    & ${
                                \begin{bmatrix}
                                  3 \times 3, 64 \\
                                  3 \times 3, 64 \\
                                \end{bmatrix}
                                \times 2
                              }$
                                                         & ${
                                                             \begin{bmatrix}
                                                               3 \times 3, 58 \\
                                                               3 \times 3, 58 \\
                                                             \end{bmatrix}
                                                             \times 2
                                                           }$
                                                                                      & ${
                                                                                          \begin{bmatrix}
                                                                                            3 \times 3, 50 \\
                                                                                            3 \times 3, 50 \\
                                                                                          \end{bmatrix}
                                                                                          \times 2
                                                                                        }$
                                                                                                                   & ${
                                                                                                                       \begin{bmatrix}
                                                                                                                         3 \times 3, 43 \\
                                                                                                                         3 \times 3, 43 \\
                                                                                                                       \end{bmatrix}
                                                                                                                       \times 2
                                                                                                                     }$
    \\
    \hline
    stage3\_x  & $\surd$    & ${
                                 \begin{bmatrix}
                                   3 \times 3, 128 \\
                                   3 \times 3, 128 \\
                                 \end{bmatrix}
                                 \times 2
                               }$
                                                         & ${
                                                             \begin{bmatrix}
                                                               3 \times 3, 116 \\
                                                               3 \times 3, 116 \\
                                                             \end{bmatrix}
                                                             \times 2
                                                           }$
                                                                                      & ${
                                                                                          \begin{bmatrix}
                                                                                            3 \times 3, 99 \\
                                                                                            3 \times 3, 99 \\
                                                                                          \end{bmatrix}
                                                                                          \times 2
                                                                                        }$
                                                                                                                   & ${
                                                                                                                       \begin{bmatrix}
                                                                                                                         3 \times 3, 86 \\
                                                                                                                         3 \times 3, 86 \\
                                                                                                                       \end{bmatrix}
                                                                                                                       \times 2
                                                                                                                     }$
    \\
    \hline
    stage4\_x  & $\surd$    & ${
                                  \begin{bmatrix}
                                    3 \times 3, 256 \\
                                    3 \times 3, 256 \\
                                  \end{bmatrix}
                                  \times 2
                                }$
                                                         & ${
                                                             \begin{bmatrix}
                                                               3 \times 3, 231 \\
                                                               3 \times 3, 231 \\
                                                             \end{bmatrix}
                                                             \times 2
                                                           }$
                                                                                      & ${
                                                                                          \begin{bmatrix}
                                                                                            3 \times 3, 197  \\
                                                                                            3 \times 3, 197  \\
                                                                                          \end{bmatrix}
                                                                                          \times 2
                                                                                        }$
                                                                                                                   & ${
                                                                                                                       \begin{bmatrix}
                                                                                                                         3 \times 3, 171  \\
                                                                                                                         3 \times 3, 171  \\
                                                                                                                       \end{bmatrix}
                                                                                                                       \times 2
                                                                                                                     }$
    \\
    \hline
    stage5\_x  & $\surd$    & ${
                                \begin{bmatrix}
                                  3 \times 3, 512 \\
                                  3 \times 3, 512 \\
                                \end{bmatrix}
                                \times 2
                              }$
                                                         & ${
                                                             \begin{bmatrix}
                                                               3 \times 3, 461 \\
                                                               3 \times 3, 461 \\
                                                             \end{bmatrix}
                                                             \times 2
                                                           }$
                                                                                      &
                                                                                                                   &
    \\
    \hline
    output\_x  & -          & \multicolumn{4}{c}{ average pool } \\
    \cline{2-6}
               &            & 1000-d fc                  & 500-d fc                   & 200-d fc                   & 100-d fc \\
    \cline{2-6}
               & -          & \multicolumn{4}{c}{ softmax } \\
    \hline
  \end{tabular}
\end{table*}

\begin{table*}[t]
  \centering
  \caption{4 stage HeteroFL architecture for image classification task on the ImageNet dataset. 
  In each client group, the number of stages is fixed at 4, 
  and the hidden channen ratio is modified so that the number of parameters become the same as the proposed ScalableFL architecture.}
  \label{tab:arch_imagenet_hetero}
  \begin{tabular}{c|c||c|c|c|c}
    \hline
    Layer name & Shared     & ImageNet-1k                & ImageNet-500               & ImageNet-200               & ImageNet-100               \\
    \hline
    stage1     & $\surd$    & 7 $\times$ 7, 128, stride 2 & 7 $\times$ 7, 115, stride 2 & 7 $\times$ 7, 50, stride 2 & 7 $\times$ 7, 43, stride 2 \\
    \hline
    stage2\_x  & -          & \multicolumn{4}{c}{ 3 $\times$ 3 max pool, stride 2 } \\
    \cline{2-6}
               & $\surd$    & ${
                                \begin{bmatrix}
                                  3 \times 3, 128 \\
                                  3 \times 3, 128 \\
                                \end{bmatrix}
                                \times 2
                              }$
                                                         & ${
                                                             \begin{bmatrix}
                                                               3 \times 3, 115 \\
                                                               3 \times 3, 115 \\
                                                             \end{bmatrix}
                                                             \times 2
                                                           }$
                                                                                      & ${
                                                                                          \begin{bmatrix}
                                                                                            3 \times 3, 50 \\
                                                                                            3 \times 3, 50 \\
                                                                                          \end{bmatrix}
                                                                                          \times 2
                                                                                        }$
                                                                                                                   & ${
                                                                                                                       \begin{bmatrix}
                                                                                                                         3 \times 3, 43 \\
                                                                                                                         3 \times 3, 43 \\
                                                                                                                       \end{bmatrix}
                                                                                                                       \times 2
                                                                                                                     }$
    \\
    \hline
    stage3\_x  & $\surd$    & ${
                                 \begin{bmatrix}
                                   3 \times 3, 256 \\
                                   3 \times 3, 256 \\
                                 \end{bmatrix}
                                 \times 2
                               }$
                                                         & ${
                                                             \begin{bmatrix}
                                                               3 \times 3, 230 \\
                                                               3 \times 3, 230 \\
                                                             \end{bmatrix}
                                                             \times 2
                                                           }$
                                                                                      & ${
                                                                                          \begin{bmatrix}
                                                                                            3 \times 3, 99 \\
                                                                                            3 \times 3, 99 \\
                                                                                          \end{bmatrix}
                                                                                          \times 2
                                                                                        }$
                                                                                                                   & ${
                                                                                                                       \begin{bmatrix}
                                                                                                                         3 \times 3, 86 \\
                                                                                                                         3 \times 3, 86 \\
                                                                                                                       \end{bmatrix}
                                                                                                                       \times 2
                                                                                                                     }$
    \\
    \hline
    stage4\_x  & $\surd$    & ${
                                  \begin{bmatrix}
                                    3 \times 3, 512 \\
                                    3 \times 3, 512 \\
                                  \end{bmatrix}
                                  \times 2
                                }$
                                                         & ${
                                                             \begin{bmatrix}
                                                               3 \times 3, 460 \\
                                                               3 \times 3, 460 \\
                                                             \end{bmatrix}
                                                             \times 2
                                                           }$
                                                                                      & ${
                                                                                          \begin{bmatrix}
                                                                                            3 \times 3, 197  \\
                                                                                            3 \times 3, 197  \\
                                                                                          \end{bmatrix}
                                                                                          \times 2
                                                                                        }$
                                                                                                                   & ${
                                                                                                                       \begin{bmatrix}
                                                                                                                         3 \times 3, 171  \\
                                                                                                                         3 \times 3, 171  \\
                                                                                                                       \end{bmatrix}
                                                                                                                       \times 2
                                                                                                                     }$
    \\
    \hline
    output\_x  & -          & \multicolumn{4}{c}{ average pool } \\
    \cline{2-6}
               &            & 1000-d fc                  & 500-d fc                   & 200-d fc                   & 100-d fc \\
    \cline{2-6}
               & -          & \multicolumn{4}{c}{ softmax } \\
    \hline
  \end{tabular}
\end{table*}

\begin{table*}[t]
  \centering
  \caption{ScalableFL architecture for object detection task on MSCOCO and PascalVOC. 
  The base model architecture is ResNet18 with two extra stages and SSD head layers. 
  Downsampling is performed by stage3\_1, stage4\_1, and stage5\_1 with a stride of 2. 
  In addition SSD head layers and all batch normalization layers are personalized layers.
  }
  \label{tab:arch_ssd}
  \begin{tabular}{c|c||c|c}
    \hline
    Layer name & Shared     & MSCOCO                       & PascalVOC \\
    \hline
    stage1     & $\surd$    & \multicolumn{2}{c}{ 7 $\times$ 7, 64, stride 2 } \\
    \hline
    stage2\_x  & -          & \multicolumn{2}{c}{ 3 $\times$ 3 max pool, stride 2 } \\
    \cline{2-4}
               & $\surd$    & \multicolumn{2}{c}{ 
                                ${
                                  \begin{bmatrix}
                                    3 \times 3, 64  \\
                                    3 \times 3, 64  \\
                                  \end{bmatrix}
                                  \times 2
                                }$
                              } \\
    \hline
    stage3\_x  & $\surd$    & \multicolumn{2}{c}{ 
                                ${
                                  \begin{bmatrix}
                                    3 \times 3, 128 \\
                                    3 \times 3, 128 \\
                                  \end{bmatrix}
                                  \times 2
                                }$
                              } \\
    \hline
    stage4\_x  & $\surd$    & \multicolumn{2}{c}{ 
                                ${
                                  \begin{bmatrix}
                                    3 \times 3, 256  \\
                                    3 \times 3, 256  \\
                                  \end{bmatrix}
                                  \times 2
                                }$
                              } \\
    \hline
    stage5\_x  & $\surd$    & \multicolumn{2}{c}{ 
                                ${
                                  \begin{bmatrix}
                                    3 \times 3, 512  \\
                                    3 \times 3, 512  \\
                                  \end{bmatrix}
                                  \times 2
                                }$
                              } \\
    \hline
    stage6\_x  & $\surd$    & 1 $\times$ 1, 128            & 1 $\times$ 1, 88              \\
    \cline{2-4}
               & $\surd$    & 3 $\times$ 3, 256, stride 2  & 3 $\times$ 3, 176, stride 2   \\
    \hline
    stage7\_x  & $\surd$    & 1 $\times$ 1, 128            &                               \\
    \cline{2-4}
               & $\surd$    & 3 $\times$ 3, 256, stride 2  &                               \\
    \hline
    head1      &            & (from stage3\_4)             & (from stage2\_4)              \\
     -class    &            & 3 $\times$ 3, 405(5 anchors) & 3 $\times$ 3, 105(5 anchors)  \\
     -location &            & 3 $\times$ 3, 20(5 anchors)  & 3 $\times$ 3, 20(5 anchors)   \\
    \hline
    head2      &            & (from stage4\_6)             & (from stage3\_4)              \\
     -class    &            & 3 $\times$ 3, 567(7 anchors) & 3 $\times$ 3, 105(7 anchors)  \\
     -location &            & 3 $\times$ 3, 28(7 anchors)  & 3 $\times$ 3, 28(7 anchors)   \\
    \hline
    head3      &            & (from stage5\_3)             & (from stage4\_6)              \\
     -class    &            & 3 $\times$ 3, 567(7 anchors) & 3 $\times$ 3, 105(7 anchors)  \\
     -location &            & 3 $\times$ 3, 28(7 anchors)  & 3 $\times$ 3, 28(7 anchors)   \\
    \hline
    head4      &            & (from stage6\_2)             & (from stage5\_3)              \\
     -class    &            & 3 $\times$ 3, 567(7 anchors) & 3 $\times$ 3, 105(7 anchors)  \\
     -location &            & 3 $\times$ 3, 28(7 anchors)  & 3 $\times$ 3, 28(7 anchors)   \\
    \hline
    head5      &            & (from stage7\_2)             & (from stage6\_2)              \\
     -class    &            & 3 $\times$ 3, 405(5 anchors) & 3 $\times$ 3, 105(5 anchors)  \\
     -location &            & 3 $\times$ 3, 20(5 anchors)  & 3 $\times$ 3, 20(5 anchors)   \\
    \hline
  \end{tabular}
\end{table*}

\begin{table*}[t]
  \centering
  \caption{7 stage HeteroFL architecture for object detection task on MSCOCO and PascalVOC. 
  In each client group, the number of stages is fixed at 7, 
  and the hidden channel ratio is modified so that the number of parameters become the same as the proposed ScalableFL architecture.
  }
  \label{tab:arch_ssd_hetero}
  \begin{tabular}{c|c||c|c}
    \hline
    Layer name & Shared     & MSCOCO                       & PascalVOC \\
    \hline
    stage1     & $\surd$    & \multicolumn{2}{c}{ 7 $\times$ 7, 64, stride 2 } \\
    \hline
    stage2\_x  & -          & \multicolumn{2}{c}{ 3 $\times$ 3 max pool, stride 2 } \\
    \cline{2-4}
               & $\surd$    & \multicolumn{2}{c}{ 
                                ${
                                  \begin{bmatrix}
                                    3 \times 3, 64  \\
                                    3 \times 3, 64  \\
                                  \end{bmatrix}
                                  \times 2
                                }$
                              } \\
    \hline
    stage3\_x  & $\surd$    & \multicolumn{2}{c}{ 
                                ${
                                  \begin{bmatrix}
                                    3 \times 3, 128 \\
                                    3 \times 3, 128 \\
                                  \end{bmatrix}
                                  \times 2
                                }$
                              } \\
    \hline
    stage4\_x  & $\surd$    & \multicolumn{2}{c}{ 
                                ${
                                  \begin{bmatrix}
                                    3 \times 3, 256  \\
                                    3 \times 3, 256  \\
                                  \end{bmatrix}
                                  \times 2
                                }$
                              } \\
    \hline
    stage5\_x  & $\surd$    & \multicolumn{2}{c}{ 
                                ${
                                  \begin{bmatrix}
                                    3 \times 3, 512  \\
                                    3 \times 3, 512  \\
                                  \end{bmatrix}
                                  \times 2
                                }$
                              } \\
    \hline
    stage6\_x  & $\surd$    & 1 $\times$ 1, 128            & 1 $\times$ 1, 61              \\
    \cline{2-4}
               & $\surd$    & 3 $\times$ 3, 256, stride 2  & 3 $\times$ 3, 123, stride 2   \\
    \hline
    stage7\_x  & $\surd$    & 1 $\times$ 1, 128            & 1 $\times$ 1, 61              \\
    \cline{2-4}
               & $\surd$    & 3 $\times$ 3, 256, stride 2  & 3 $\times$ 3, 123, stride 2   \\
    \hline
    head1      &            & (from stage3\_4)             & (from stage3\_4)              \\
     -class    &            & 3 $\times$ 3, 405(5 anchors) & 3 $\times$ 3, 105(5 anchors)  \\
     -location &            & 3 $\times$ 3, 20(5 anchors)  & 3 $\times$ 3, 20(5 anchors)   \\
    \hline
    head2      &            & (from stage4\_6)             & (from stage4\_6)              \\
     -class    &            & 3 $\times$ 3, 567(7 anchors) & 3 $\times$ 3, 105(7 anchors)  \\
     -location &            & 3 $\times$ 3, 28(7 anchors)  & 3 $\times$ 3, 28(7 anchors)   \\
    \hline
    head3      &            & (from stage5\_3)             & (from stage5\_3)              \\
     -class    &            & 3 $\times$ 3, 567(7 anchors) & 3 $\times$ 3, 105(7 anchors)  \\
     -location &            & 3 $\times$ 3, 28(7 anchors)  & 3 $\times$ 3, 28(7 anchors)   \\
    \hline
    head4      &            & (from stage6\_2)             & (from stage6\_2)              \\
     -class    &            & 3 $\times$ 3, 567(7 anchors) & 3 $\times$ 3, 105(7 anchors)  \\
     -location &            & 3 $\times$ 3, 28(7 anchors)  & 3 $\times$ 3, 28(7 anchors)   \\
    \hline
    head5      &            & (from stage7\_2)             & (from stage7\_2)              \\
     -class    &            & 3 $\times$ 3, 405(5 anchors) & 3 $\times$ 3, 105(5 anchors)  \\
     -location &            & 3 $\times$ 3, 20(5 anchors)  & 3 $\times$ 3, 20(5 anchors)   \\
    \hline
  \end{tabular}
\end{table*}

\clearpage
\begin{center}
{\bf Appendix B: Learning Condition Details} \\
\end{center}
In Table \ref{tab:ex_imagenet} and Table \ref{tab:ex_ssd}, we describe the details of the experimental settings.

\begin{table*}[t]
  \centering
  \caption{Experimental settings for image classification task on ImageNet dataset. }
  \label{tab:ex_imagenet}
  \begin{tabular}{ll|rrrr}
    \hline
                  &                            & ImageNet-1k            & ImageNet-500           & ImageNet-200           & ImageNet-100           \\
    \hline
    Data          & Input image size [pixel]   & 256                    & 192                    & 128                    & 96                     \\
                  & \# of classes              & 1,000                  & 500                    & 200                    & 100                    \\
                  & \# of clients              & 20                     & 20                     & 20                     & 20                     \\
                  & \# of training images      & 16k                    & 16k                    & 16k                    & 16k                    \\
                  & \# of test images          & 50k                    & 25k                    & 10k                    & 5k                     \\
    \hline
    \hline
    Model         & \# of stages               & 5                      & 5                      & 4                      & 4                      \\
    (ScalableFL \&& Hidden channel ratio       & 1.0                    & 0.9                    & 0.77                   & 0.67                   \\
     Individual)  & \# of parameters           & 11.7M                  & 9.31M                  & 1.70M                  & 1.26M                 \\
    \hline
    Model         & \# of stages               & 4                      & 4                      & 4                      & 4                      \\
    (HeteroFL     & Hidden channel ratio       & 2.0                    & 1.8                    & 0.77                   & 0.67                   \\
     4 stages)    & \# of parameters           & 11.6M                  & 9.25M                  & 1.70M                  & 1.26M                  \\
    \hline
    Model         & \# of stages               & 5                      & 5                      & 5                      & 5                      \\
    (HeteroFL     & Hidden channel ratio       & 1.0                    & 0.9                    & 0.38                   & 0.33                   \\
     5 stages)    & \# of parameters           & 11.7M                  & 9.31M                  & 1.67M                  & 1.25M                  \\
    \hline
    \hline
    Learning      & \# of communication rounds & \multicolumn{4}{c}{ 5k                                                                          } \\
    (ScalableFL \&& \# of local iterations     & \multicolumn{4}{c}{ 10                                                                          } \\
     HeteroFL)    & participation rate [\%]    & \multicolumn{4}{c}{ 10                                                                          } \\
                  & Local optimizer            & \multicolumn{4}{c}{ SGD (momentum = 0.9, Nesterov = False)                                      } \\
                  & Learning rate              & \multicolumn{4}{c}{ initially 0.2, cosine decay schedule                                        } \\
                  & Weight decay               & \multicolumn{4}{c}{ 5e-4 (only for weight of conv/linear)                                       } \\
                  & Mini batch size            & 256 (8 GPUs)           & 256 (8 GPUs)           & 256 (8 GPUs)           & 256 (8 GPUs)           \\
    \hline
    Learning      & \# of iterations           & \multicolumn{4}{c}{ 5k / 50k / 100k                                                             } \\
    (Individual)  & Optimizer                  & \multicolumn{4}{c}{ SGD (momentum = 0.9, Nesterov = False)                                      } \\
                  & Initial learning rate      & 0.05                   & 0.05                   & 0.05                   & 0.05                   \\
                  & Learning rate schedule     & \multicolumn{4}{c}{ Cosign decay schedule                                                       } \\
                  & Weight decay               & 5e-2                   & 5e-2                   & 5e-2                   & 5e-2                   \\
                  & Mini batch size            & 256 (8 GPUs)           & 256 (8 GPUs)           & 256 (8 GPUs)           & 256 (8 GPUs)           \\
    \hline
    \hline
    Preprocessing & Random resized crop [pixel]& 256                    & 192                    & 128                    & 96                     \\
    (training)    & Random hflip [\%]          & \multicolumn{4}{c}{ 50                                                                          } \\
    \hline
    Preprocessing & resize [pixel]             & 293                    & 219                    & 146                    & 110                    \\
    (validation)  & center crop [pixel]        & 256                    & 192                    & 128                    & 96                     \\
    \hline
    Normalization & Mean [R, G, B]             & \multicolumn{4}{c}{ [ 0.485, 0.456, 0.406 ]                                                     } \\
                  & SD [R, G, B ]              & \multicolumn{4}{c}{ [ 0.229, 0.224, 0.225 ]                                                     } \\
    \hline
  \end{tabular}
\end{table*}

\begin{table*}[t]
  \centering
  \caption{Experimental settings for object detection task on MSCOCO and PascalVOC.}
  \label{tab:ex_ssd}
  \begin{tabular}{ll|rr}
    \hline
                  &                              & MSCOCO                 & PascalVOC                                       \\
    \hline
    Data          & \# of classes                & 80                     & 20                                              \\
                  & Input image size [pixel]     & 512                    & 256                                             \\
                  & \# of training images        & 7,329                  & 8,275                                           \\
                  & \# of test images            & 5,000                  & 4,952                                           \\
    \hline
    \hline
    Model         & \# of stages                 & 7                      & 6                                               \\
    (ScalableFL \&& Hidden channel ratio         & 1.0                    & 0.68                                            \\
     Individual)  & \# of parameters             & 18.8M                  & 13.0M                                           \\
    \hline
    Model         & \# of stages                 & 6                      & 6                                               \\
    (HeteroFL   \&& Hidden channel ratio         & 1.7                    & 0.68                                            \\
     6 stages)    & \# of parameters             & 18.9M                  & 13.0M                                           \\
    \hline
    Model         & \# of stages                 & 7                      & 7                                               \\
    (HeteroFL   \&& Hidden channel ratio         & 1.0                    & 0.48                                            \\
     7 stages)    & \# of parameters             & 18.8M                  & 13.0M                                           \\
    \hline
    \hline
    Learning      & \# of communication rounds   & \multicolumn{2}{c}{ 2000                                               } \\
    (ScalableFL \&& \# of local iterations       & \multicolumn{2}{c}{ 25                                                 } \\
     HeteroFL)    & participation rate [\%]      & \multicolumn{2}{c}{ 20                                                 } \\
                  & Local optimizer              & \multicolumn{2}{c}{ SGD (momentum = 0.9, Nesterov = False)             } \\
                  & Initial learning rate        & \multicolumn{2}{c}{ 1e-2                                               } \\
                  & Learning rate warmup         & \multicolumn{2}{c}{ during 4[\%] communication rounds                  } \\
                  & Learning rate decay          & \multicolumn{2}{c}{ $\times$ 0.1 after 70[\%] \& 90[\%] communication rounds } \\
                  & Weight decay                 & \multicolumn{2}{c}{ 5e-4 (only for weight of conv/fc)                  } \\
                  & Mini batch size              & 64 (4 GPUs)            & 64 (4 GPUs)                                     \\
    \hline
    Learning      & \# of iterations             & \multicolumn{2}{c}{ 5k / 50k / 100k                                    } \\
    (Individual)  & Optimizer                    & \multicolumn{2}{c}{ SGD (momentum = 0.9, Nesterov = False)             } \\
                  & Initial learning rate        & 1e-2 / 1e-1 / 1e-1     & 1e-2 / 1e-2 / 1e-2                              \\
                  & Learning rate warmup         & \multicolumn{2}{c}{ during 4[\%] trainig iterations                    } \\
                  & Learning rate decay          & \multicolumn{2}{c}{ $\times$ 0.1 after 70[\%] \& 90[\%] trainig iterations } \\
                  & Weight decay                 & 1e-3 / 1e-5 / 1e-5     & 1e-3 / 1e-3 / 1e-3                              \\
                  & Mini batch size              & 64 (2 GPUs)            & 64 (2 GPUs)                                     \\
    \hline
    \hline
    Preprocessing & Random expand                & \multicolumn{2}{c}{ max ratio = 3.0                                    } \\
    (training)    & Random sample                & \multicolumn{2}{c}{ min ratio = 0.3                                    } \\
                  & Random hflip [\%]            & \multicolumn{2}{c}{ 50                                                 } \\
                  & Resize [pixel]               & 512                    & 256                                             \\
                  & Color jitter                 & \multicolumn{2}{c}{ brightness = 32/256, contrast = 0.5, saturation = 0.5, hue = 25.4 } \\
    \hline
    Preprocessing & Resize [pixel]               & 512                    & 256                                             \\
    (evaluation)  &                              &                        &                                                 \\
    \hline
    Normalization & Mean [R, G, B]               & \multicolumn{2}{c}{ [ 0.485, 0.456, 0.406 ]                            } \\
                  & SD [R, G, B ]                & \multicolumn{2}{c}{ [ 0.229, 0.224, 0.225 ]                            } \\
    \hline
    Evaluation    & Metric                       & MSCOCO                 & VOC '07                                         \\
                  &                              & (IoU=[0.5,0.95])       &                                                 \\
                  & IoU threshold for metric     & \multicolumn{2}{c}{ 0.5                                                } \\
                  & Probability threshold        & \multicolumn{2}{c}{ 0.03                                               } \\
                  & IoU threshold for NMS        & \multicolumn{2}{c}{ 0.45                                               } \\
                  & Top k value for NMS          & \multicolumn{2}{c}{ 200                                                } \\
    \hline
  \end{tabular}
\end{table*}

\clearpage
\begin{center}
{\bf Appendix C: Proofs of Theories} \\
\end{center}
Remember that $\bar{n}_j := \sum_{m'=j}^M n_{m'}$ and let $n = \sum_{m=1}^M n_{m}$.
First, notice that 
\begin{align*}
\hat{L}(f) - L(f)
& = 
\sum_{m=1}^M \alpha_m (\hat{L}_m(f^{(m)}) - L_m(f^{(m)})) \\
& =
\sum_{m=1}^M \alpha_m \sum_{j=1}^{m} \left(\hat{L}_m(f^{(j)}) - L_m(f^{(j)}) - \hat{L}_m(f^{(j-1)}) + L_m(f^{(j-1)})\right) \\
& =
\sum_{j=1}^M   \sum_{m=j}^{M} \alpha_m \left(\hat{L}_m(f^{(j)}) - L_m(f^{(j)}) - \hat{L}_m(f^{(j-1)}) + L_m(f^{(j-1)})\right).
\end{align*}

Let 
the weighted average of distribution $P_m$ be $\bar{P} := \sum_{m=1}^M \alpha_m P_m$. 
We also introduce the following norms: 
$$
\|f\|_n^2 := \frac{1}{n}\sum_{m=1}^M \sum_{i=1}^{n_m} f(z_i^{(m)})^2,~~~\|f\|_{L^2(\bar{P})}^2 := \sum_{m=1}^M \alpha_m \|f\|^2_{L^2(P_m)}.
$$

The local Rademacher complexity is characterized by the population $L_2$-norm.
However, our assumption only asserts the boundedness of the empirical $L_2$-norm.
To bridge this gap, we need to 
bound the population $\LPi$-distance $\|\hat{f}^{(j)} - \hat{f}^{(j-1)}\|_{\LPi(\bar{P}_j)}$ in terms of the empirical $L_2$-norm 
$\|\fhat^{(j)} - \fhat^{(j-1)}\|_{\bar{n}_j} = 
\sqrt{\frac{1}{\bar{n}_j}\sum_{m=j}^M n_m \|\fhat^{(j)} - \fhat^{(j-1)}\|_{n_m}^2} \leq \hat{r}_j$.
To do so, we also utilize the local Rademacher complexity $\bar{R}_{j,r}(\calF_j)$.
From the assumption, there exists a function $\phi_j:(0,\infty) \to [0,\infty)$ such that 
$$
\bar{R}_{j,r}(\calF_j) \leq \phi_j(r)
$$
and 
$$
\phi_j(2 r) \leq 2 \phi_j(r).
$$
Then, by the so-called {\it peeling device}, we can show that  for any $r > 0$, 
\begin{align*}
P\left( \sup_{h = f^{(j)} - f^{(j-1)}: f^{(j)}\in \calF_j} \frac{\|h\|_{\LPi(\bar{P}_j)}^{2} -  \|h\|_{\bar{n}_j}^{2}}{\|h\|_{\LPi(\bar{P}_j)}^{2}+ {r}^2}
\geq 8 \frac{\phi_j(r)}{r^2} + B \sqrt{\frac{4t}{r^2 \bar{n}_j}} + B^2 \frac{2t}{r^2 \bar{n}_j} \right)
\leq e^{-t}
\end{align*}
for all $t > 0$ 
(Theorem 7.7 and Eq. (7.17) of \cite{Book:Steinwart:2008}).
Hence, if we choose $r^*_j = r^*_j(t)$ so that 
$$
8 \frac{\phi(r^*_j)}{r_j^{*2}} + B \sqrt{\frac{4t}{r_j^{*2} \bar{n}_j}} + B^2 \frac{2t}{r_j^{*2} \bar{n}_j} \leq \frac{1}{2},
$$
then it holds that 
$$
\|h\|_{\LPi(\bar{P}_j)}^{2} \leq 2 \|h\|_{\bar{n}_j}^{2} + 2 r_j^{*2}
$$
uniformly over all $h =  f^{(j)} - f^{(j-1)}$ for $f^{(j)}\in \calF_j$ with probability greater than $1-e^{-t}$.
We let this event as $\calE_{1,j}(t)$.
In this event, if $\|\fhat^{(j)} - \fhat^{(j-1)}\|_{n_m}^2 \leq \hat{r}_j^2~(\forall m \geq j)$, then it holds that
$$
\|\fhat^{(j)} - \fhat^{(j-1)}\|_{\LPi(\bar{P}_j)}^2 \leq 2 (\hat{r}_j^2 + r_j^{*2}) = \dot{r}_j^2.
$$

Now, we write $ \calF_j(r) := \{f^{(j)} \in \calF_j \mid \|\fhat^{(j)} - \fhat^{(j-1)}\|_{\LPi(\bar{P}_j)} \leq r \}$. 
Then, in the event $\calE_{1,j}(t)$, we have that 
\begin{align*}
& \frac{1}{\sum_{m=j}^{M} \alpha_m} \sum_{m=j}^{M} \alpha_m \left(\hat{L}_m(f^{(j)}) - L_m(f^{(j)}) - \hat{L}_m(f^{(j-1)}) + L_m(f^{(j-1)})\right) \\
&\leq  
\sup_{f^{(j)} \in \calF_j(\dot{r}_j)}
\frac{1}{\sum_{m=j}^{M} \alpha_m} \sum_{m=j}^{M} \alpha_m \left(\hat{L}_m(f^{(j)}) - L_m(f^{(j)}) - \hat{L}_m(f^{(j-1)}) + L_m(f^{(j-1)})\right).
\end{align*}

To bound this term, we apply the Talagrand's concentration inequality (Proposition \ref{prop:TalagrandConcent} and \cite{Talagrand2,BousquetBenett}).
To apply it, we should bound the variance and $L^\infty$-norm of 
$\ell(f^{(j)},z) - \ell(f^{(j-1)},z) - (\EE[\ell(f^{(j)},Z^{(m)})] - \EE[\ell(f^{(j-1)},Z^{(m)})])$
for any $f^{(j)} \in \calF_j(r)$ (where $r$ will be set $2 (\hat{r}_j^2 + r_j^{*2})$).
Due to the Lipschitz continuity of $\ell$, we have that 
$$
\Var[\ell(f^{(j)},Z^{(m)}) - \ell(f^{(j-1)},Z^{(m)})] \leq \Var[f^{(j)}(X^{(m)}) - f^{(j-1)}(X^{(m)})] \leq \|f^{(j)}-f^{(j-1)}\|_{\LPi(P_m)}^2.
$$
Therefore, we have that 
\begin{align*}
\frac{1}{\bar{n}_j} \sum_{m=j}^M\sum_{i=1}^{n_m} \Var[\ell(f^{(j)},Z_i^{(m)}) - \ell(f^{(j-1)},Z_i^{(m)})]
& \leq \frac{1}{\bar{n}_j} \sum_{m=j}^M\sum_{i=1}^{n_m}  \|f^{(j)}-f^{(j-1)}\|_{\LPi(P_m)}^2 \\
& =\|f^{(j)}-f^{(j-1)}\|_{\LPi(\bar{P}_j)}^2 \leq \dot{r}_j^2.
\end{align*}
Similarly, it holds that 
\begin{align*}
& |\ell(f^{(j)},z) - \ell(f^{(j-1)},z) - (\EE[\ell(f^{(j)},Z^{(m)})] - \EE[\ell(f^{(j-1)},Z^{(m)})])| \\
& \leq 
|\ell(f^{(j)},z)  - \EE[\ell(f^{(j)},Z^{(m)})]| + | \ell(f^{(j-1)},z)  -  \EE[\ell(f^{(j-1)},Z^{(m)})])|
\leq 2 B.
\end{align*}
Hence, by the Talagrand's concentration inequality (Proposition \ref{prop:TalagrandConcent} and \cite{Talagrand2,BousquetBenett}), 
for 
\begin{align*}
& \Phi_j(\dot{r}_j) := \sup_{f^{(j)} \in \calF_j(\dot{r}_j)}
\frac{1}{\sum_{m=j}^{M} \alpha_m} \sum_{m=j}^{M} \alpha_m \left(\hat{L}_m(f^{(j)}) - L_m(f^{(j)}) - \hat{L}_m(f^{(j-1)}) + L_m(f^{(j-1)})\right),
\end{align*}
it holds that
\begin{align*}
& \Phi_j(\dot{r}_j) 
\leq 
2 \EE\left[\Phi_j(\dot{r}_j) \right]
+
\dot{r}_j \sqrt{\frac{2 t}{\bar{n}_j}}
+
\frac{4 t B}{\bar{n}_j},
\end{align*}
with probability at least $1 - e^{-t}$ for any $t > 0$.
By the symmetrization argument (see Lemma 11.4 of \cite{boucheron2013concentration} for example), 
the first term in the right hand side can be bounded as 
\begin{align*}
& \EE\left[  \Phi_j(\dot{r}_j) \right] 
\leq 2 \EE_{D_n,\epsilon}\left[\sup_{f^{(j)} \in \calF_j(\dot{r}_j)} 
\frac{1}{\bar{n}_j} \sum_{m=j}^M \sum_{i=1}^{n_m} \epsilon_{i,m}
( \ell(f^{(j)},z_i^{(m)}) - \ell(f^{(j-1)},z_i^{(m)}))\right].
\end{align*}
Using $\dot{r}_{j}$, let $\hat{\gamma}_n = \hat{\gamma}_n(D_n) := \sup\{\|f^{(j)} - f^{(j-1)}\|_{\bar{n}_j}\mid \|f^{(j)} - f^{(j-1)}\|_{\LPi(\bar{P}_j)} \leq \dot{r}_{j}, f^{(j)} \in \calF_j \}$. 
It is known that the integrant of the right hand side can be bounded by a constant times the following {\it Dudley integral} (see Theorem 5.22 of \cite{wainwright2019high} or Lemma A.5 of \cite{bartlett2017spectrally:arXiv} for example):
\begin{align}
& 
\sup_{f^{(j)} \in \calF_j(\dot{r}_j)} 
\frac{1}{\bar{n}_j} \sum_{m=j}^M \sum_{i=1}^{n_m} \epsilon_{i,m}
[ \ell(f^{(j)},z_i^{(m)}) - \ell(f^{(j-1)},z_i^{(m)})] \notag \\
& \leq  \inf_{\alpha > 0}\left[\alpha + \int_\alpha^{\hat{\gamma}_n} 
\sqrt{\frac{\log(\calN(\{\ell(f^{(j)},\cdot) - \ell(f^{(j-1)},\cdot)\mid f \in \calF, \|f^{(j)} - f^{(j-1)}\|_{\LPi(\bar{P}_j)} \leq\dot{r}_j \}, \|\cdot\|_{\bar{n}_j}, \epsilon))}{\bar{n}_j}} \dd \epsilon \right] \notag \\
& 
\leq \frac{1}{\bar{n}_j} + \int_{1/\bar{n}_j}^{\hat{\gamma}_n} 
\sqrt{\frac{\log(\calN(\{\ell(f^{(j)},\cdot) - \ell(f^{(j-1)},\cdot)\mid f \in \calF, \|f^{(j)} - f^{(j-1)}\|_{\LPi(\bar{P}_j)} \leq \dot{r}_j \}, \|\cdot\|_{\bar{n}_j}, \epsilon))}{\bar{n}_j}} \dd \epsilon,
\label{eq:DudleyBoundFirst}
\end{align}
where $\calN(\cdot)$ represents the covering number of the model (Definition \ref{def:CoveringNumber}). 
Note that $|\ell(f,z)  - \ell(g,z)| \leq |f(x)-g(x)|$ by the Lipschitz continuity of $\ell$.
Therefore, the covering number of $\ell(f^{(j)},\cdot) - \ell(f^{(j-1)},\cdot)$ satisfies
\begin{align*}
& \calN(\{\ell(f^{(j)},\cdot) - \ell(f^{(j-1)},\cdot)\mid f \in \calF, \|f^{(j)} - f^{(j-1)}\|_{\LPi(\bar{P}_j)} \leq r \}, \|\cdot\|_{\bar{n}_j}, \epsilon) \\
& \leq
\calN(\{f^{(j)} -f^{(j-1)} \mid f \in \calF, \|f^{(j)} - f^{(j-1)}\|_{\LPi(\bar{P}_j)} \leq r \}, 
\|\cdot\|_{\bar{n}_j}, \epsilon).
\end{align*} 
Then, the Sudakov's minoration (Corollary 4.14 of \cite{Book:Ledoux+Talagrand:1991}) gives an upper bound of the right hand side of the second term in \Eqref{eq:DudleyBoundFirst}:
\begin{align*}
& 
\int_{1/\bar{n}_j}^{\hat{\gamma}_n} 
\sqrt{\frac{\log(\calN(\{\ell(f^{(j)},\cdot) - \ell(f^{(j-1)},\cdot)\mid f \in \calF, \|f^{(j)} - f^{(j-1)}\|_{\LPi(\bar{P}_j)} \leq r \}, \|\cdot\|_{\bar{n}_j}, \epsilon))}{\bar{n}_j}} \dd \epsilon \\
& \leq
\int_{1/\bar{n}_j}^{\hat{\gamma}_n} 
\sqrt{\frac{\log(\calN(\{f^{(j)}- f^{(j-1)} \mid f \in \calF, \|f^{(j)} - f^{(j-1)}\|_{\LPi(\bar{P}_j)} \leq r \}, \|\cdot\|_{\bar{n}_j}, \epsilon))}{\bar{n}_j}} \dd \epsilon \\
& \leq 
 \int_{1/\bar{n}_j}^{\hat{\gamma}_n} 
 \hat{R}_{j,r}(\calF_j)\sqrt{\log(\bar{n}_j)}  \frac{1}{\epsilon} \dd \epsilon 
\leq 
 \hat{R}_{j,r}(\calF_j) \sqrt{\log(\bar{n}_j)} \log(\bar{n}_j \hat{\gamma}_n),
\end{align*}
where $\hat{R}_{j,r}(\calF_j) :=\EE_{\epsilon}\left[\sup\{\frac{1}{\bar{n}_j} \sum_{m=j}^M 
\sum_{i=1}^{n_j} \epsilon_{i,m}(f^{(j)}(x_i^{(m)}))-f^{(j-1)}(x_i^{(m)}))
\mid f^{(j)} \in\calF_j, \|f^{(j)}-f^{(j-1)}\|_{\LPi(\bar{P}_j)} \leq r \} \right]$.
Since $\hat{\gamma}_n \leq 2B$, 
the expectation of the right hand side with respect to $D_n$ is 
$
 \bar{R}_{j, r}(\calF_j) \sqrt{\log(n)} \log(2 n B).
$
Therefore, we have that 
$$
\EE\left[  \Phi_j(\dot{r}_j) \right] \leq C\left( \frac{1}{\bar{n}_j} + \bar{R}_{j, \dot{r}_j}(\calF_j) \sqrt{\log(\bar{n}_j)} \log(2 \bar{n}_j B) \right),
$$
for a universal constant $C>0$ with probability at least $1 -e^{-t}$ for all $t > 0$.
We denote by this event as $\calE_{2,j}$.

Combining these inequalities, 
\begin{align*}
& \frac{1}{\sum_{m=j}^{M} \alpha_m} \sum_{m=j}^{M} \alpha_m \left(\hat{L}_m(f^{(j)}) - L_m(f^{(j)}) - \hat{L}_m(f^{(j-1)}) + L_m(f^{(j-1)})\right) \\
& \leq 
C \left[ \bar{R}_{j, \dot{r}_j}(\calF_j) \sqrt{\log(n)} \log(2 \bar{n}_j B)  + \dot{r}_j \sqrt{\frac{t}{\bar{n}_j}} + \frac{1 + t B }{\bar{n}_j}\right],
\end{align*}
holds in the event $\calE_{1,j}(t) \cap \calE_{2,j}(t)$.
Taking the uniform bound for all $j = 1,\dots,M$, we obtain that 
$$
\hat{L}(\fhat) - L(\fhat)
\leq C \sum_{j=1}^M \left(\sum_{m=j}^M \alpha_m\right)
 \left[ \bar{R}_{j, \dot{r}_j}(\calF_j) \sqrt{\log(n)} \log(2 \bar{n}_j B)  + \dot{r}_j \sqrt{\frac{t + \log(2M)}{\bar{n}_j}} + \frac{1 + B( t  +\log(2M)) }{\bar{n}_j}\right],
$$
with probability $1 - \exp(-t)$, where we redefine $t \leftarrow t + \log(2M)$.

\vspace{1cm}
\begin{center}
{\bf Appendix D: Auxiliary Lemmas} \\
\end{center}
\begin{Definition}[Covering number]\label{def:CoveringNumber}
For a metric space $\tilde{\calF}$ equipped with a metric $\tilde{d}$, the $\epsilon$-covering number $\calN(\tilde{\calF},\tilde{d},\epsilon)$
is defined as the minimum number of balls with radius $\epsilon$ (measured by the metric $\tilde{d}$) to cover the metric space $\tilde{\calF}$. 
\end{Definition}

\begin{Proposition}[Talagrand's Concentration Inequality adapted \cite{Talagrand2,BousquetBenett}]
\label{prop:TalagrandConcent}
Let $\calG$ be a function class on $\calX$ that is separable with respect to $\infty$-norm, and 
$\{x_i\}_{i=1}^n$ be independent random variables with values in $\calX$.
Furthermore, let $\sigma^2 \geq 0$ and $U\geq 0$ be 
$\sigma^2 := \frac{1}{n}\sum_{i=1}^n \sup_{g \in \calG} \EE[(g(X_i)-\EE[g(X_i)])^2]$ and $U := \sup_{g \in \calG} \|g\|_{\infty}$,
then for $Z := \sup_{g\in \calG}\left|\frac{1}{n} \sum_{i=1}^n g(x_i) - \EE[g] \right|$, we have
\begin{align}
P\left( Z \geq 2 \EE[Z] + \sqrt{\frac{2 \sigma^2 t}{n}} + \frac{2 U t}{n}  \right) \leq e^{-t},
\label{eq:TalagrandIneq}
\end{align}
for all $t >  0$.
\end{Proposition}

\end{document}